\documentclass{article}

\usepackage{microtype}
\usepackage{graphicx}
\usepackage{subfigure}
\usepackage{booktabs} 
\graphicspath{ {figures/} }
\renewcommand{\cite}[1]{\citep{#1}}
\usepackage{diagbox}
\usepackage{makecell}
\usepackage{float}
\usepackage{dblfloatfix}

\usepackage[accepted]{icml2023_arxiv}

\usepackage{amsmath}
\usepackage{amssymb}
\usepackage{mathtools}
\usepackage{amsthm}

\usepackage[capitalize,noabbrev]{cleveref}

\theoremstyle{plain}

\theoremstyle{definition}

\theoremstyle{remark}

\usepackage[textsize=tiny]{todonotes}

\icmltitlerunning{Naive Few-Shot Learning}

\begin{document}

\twocolumn[
\icmltitle{Naive Few-Shot Learning:\\Uncovering the fluid intelligence of machines}

\icmlsetsymbol{equal}{*}

\begin{icmlauthorlist}
\icmlauthor{Tomer Barak}{xxx}
\icmlauthor{Yonatan Loewenstein}{xxx,yyy}

\end{icmlauthorlist}

\icmlaffiliation{xxx}{The Edmond and Lily Safra Center for Brain Sciences, The Hebrew University, Jerusalem}
\icmlaffiliation{yyy}{Department of Cognitive Sciences, The Federmann Center for the Study of Rationality, The Alexander Silberman Institute of Life Sciences, The Hebrew University, Jerusalem}

\icmlcorrespondingauthor{Tomer Barak}{tomer.barak@mail.huji.ac.il}

\icmlkeywords{Machine Learning, ICML, Fluid intelligence, Consistency evaluation, few-shot learning, naive few-shot learning}

\vskip 0.3in
]



\printAffiliationsAndNotice{}  

\begin{abstract}
In this paper, we aimed to help bridge the gap between human fluid intelligence - the ability to solve novel tasks without prior training - and the performance of deep neural networks, which typically require extensive prior training. An essential cognitive component for solving intelligence tests, which in humans are used to measure fluid intelligence, is the ability to identify regularities in sequences. This motivated us to construct a benchmark task, which we term \textit{sequence consistency evaluation} (SCE), whose solution requires the ability to identify regularities in sequences. Given the proven capabilities of deep networks, their ability to solve such tasks after extensive training is expected. Surprisingly, however, we show that naive (randomly initialized) deep learning models that are trained on a \textit{single} SCE with a \textit{single} optimization step can still solve non-trivial versions of the task relatively well. We extend our findings to solve, without any prior training, real-world anomaly detection tasks in the visual and auditory modalities. These results demonstrate the fluid-intelligent computational capabilities of deep networks. We discuss the implications of our work for constructing fluid-intelligent machines.
\end{abstract}

\section{Introduction}
\label{introduction}


It has been demonstrated that deep learning models can successfully solve intelligence tests \cite{santoro_simple_2017, barrett_measuring_2018, zhuo_solving_2020, kim_few-shot_2020, Webb2022EmergentAR}. However, they required extensive prior training to achieve this goal. Are deep learning models capable of exhibiting ``fluid intelligence'' that does not rely on prior training?

To address this question, we focused on a specific sub-task shared among presumably different intelligence tests, the extraction of simple rules from sequences, \cite{sternberg_component_1977, sternberg_components_1983, lohman_complex_2000, siebers_computer_2015}. For example, when solving a Raven's Progression Matrix \cite{raven_manual_1998}, humans extract the rules governing the change in the matrix's rows and columns, and then use these rules to select the most consistent answer \cite{carpenter_what_1990}. Our focus here is the extent to which \textit{naive} models can solve this computational task, extracting simple rules from sequences of inputs \textit{without prior training}. This approach is an unusual setting for deep learning models, in which pretraining is considered crucial even in the context of few-shot learning \cite{chollet_measure_2019, vogelstein_prospective_2022}.

We will start by describing the SCE, a task designed to study rule extraction. Then we will introduce Contrastive Predictive Coding (CPC) and Relation Network (RN) models and compare their performances on the SCE task. Our main result is that CPC can successfully solve non-trivial versions of the task by a single optimization step (starting with random weights) over a single sequence of $5$ images. We conclude by demonstrating the applicability of our approach to real-world problems using two anomaly detection tasks.

\begin{figure*}[t]
\begin{center}
\centerline{\includegraphics[width=\linewidth]{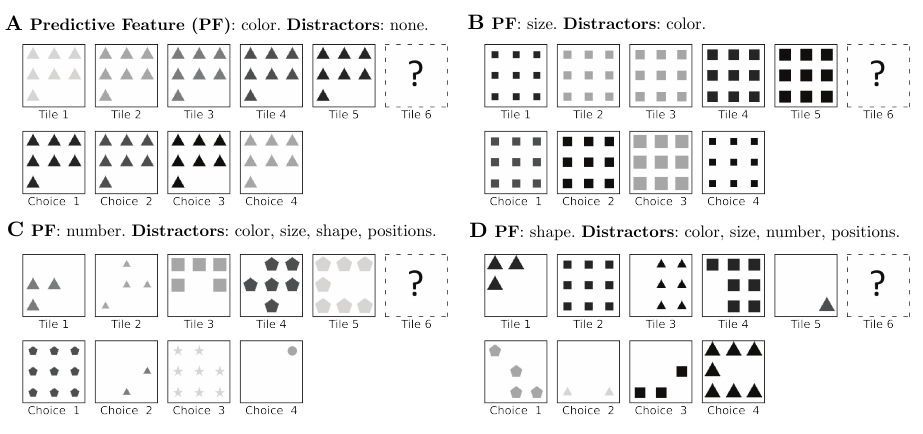}}
\caption{\textbf{SCE Tests.} The predictive features can be the (A) Color, (B) Size, (C) Number, which increases monotonically, or the (D) Shape of the objects, which alternates between a triangle and a square. Given a predictive feature, the rest of the features are either constant or random. We refer to the random features as \textit{distractors}, and their number determines the test difficulty. \textit{The correct choice in all of the tests above is $3$}.}
\label{fig:sequences_examples}
\end{center}
\vskip -0.2in
\end{figure*}

\begin{figure*}[t]
\begin{center}
\centerline{\includegraphics[width=\linewidth]{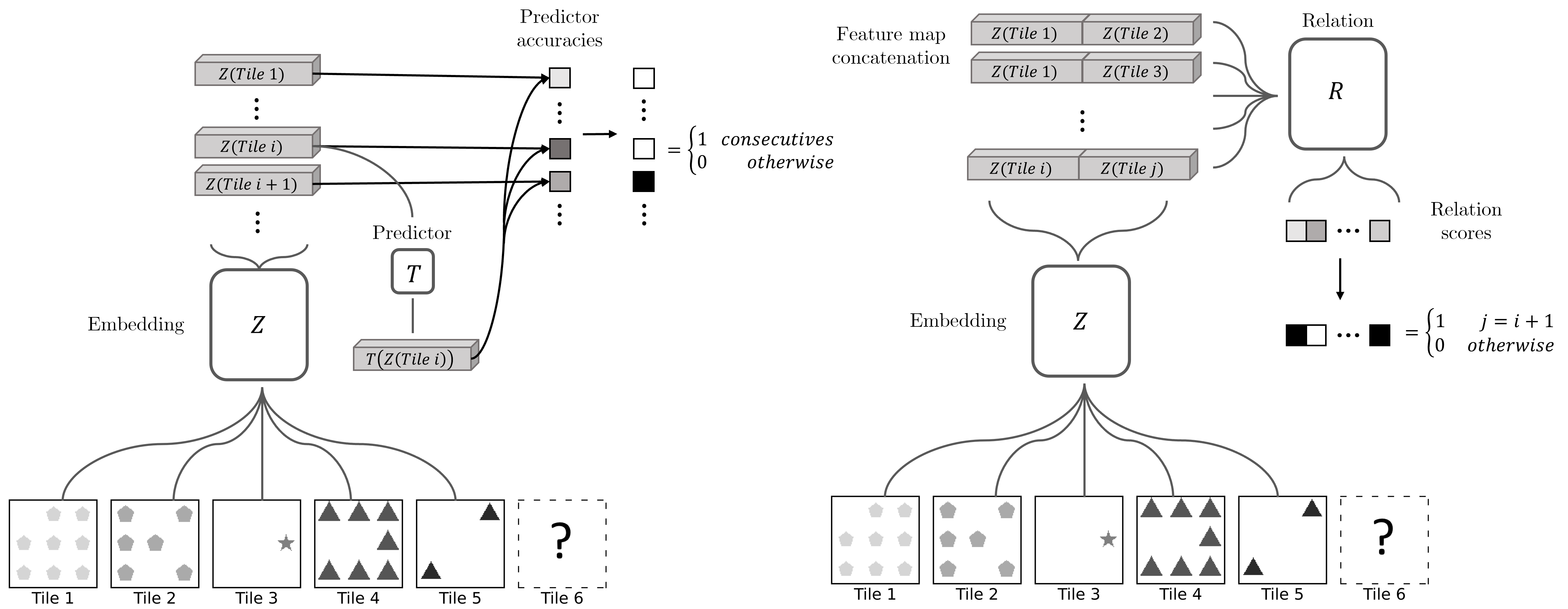}}
\caption{\textbf{Comparison between Markov-CPC (left) and RN (right).} The main difference between the models is that Markov-CPC imposes a \textit{causal structure} between consecutive latent representations, while RN can learn a \textit{general relation} $R$ between two latent representations.}
\label{fig:two_models}
\end{center}
\end{figure*}

\section{Methods}
\subsection{Sequence consistency evaluation (SCE) tests}
\label{sec:data}

Each SCE test\footnote{Code is available in https://github.com/Tomer-Barak/Naive-Few-Shot-Learning} is a sequence of $K=5$ gray-scale images $\mathbf{x}_j$ and $n=4$ optional-choice images (Fig. \ref{fig:sequences_examples}). Each image includes 1-9 identical objects arranged on a 3$\times$3 grid. An image is characterized by a low-dimensional vector of features, $\mathbf{f}_j$, where $f^i_j$ denotes the value of feature $i$ in image $j$. We use the following five features: the number of objects in an image (possible values: 1 to 9), their color (6 linearly distributed gray-scale values), the shapes (circle, triangle, square, star, hexagon), their size (6 linearly distributed values for the shapes' enclosing circle circumference), and positions (a vector of grid positions that was used to place the shapes in order). An image $\mathbf{x}_j$ is constructed according to its characterizing features by a non-linear and complex generative function $\mathbf{x}_j=G\left(\mathbf{f}_j\right)$.

One of the features $f^p$ predictably changes along the sequence according to a simple deterministic rule $f_{j+1}^p=U(f_j^p)$ while the other features are either constant over the images or change randomly (values are i.i.d). We refer to the randomly-changing features as \emph{distractors}, and their number is considered a measure of the difficulty of the test. Given a sequence of $K$ images, an agent's task is to select the correct $K+1^{\text{th}}$ image from the set of $n$ optional choice images that are generated using the same generative function $G$ from the feature space. In the correct choice, $f^p$ follows the deterministic rule $f_{K+1}^p=U(f_K^p)$, whereas in the incorrect choices it does not follow that rule and is instead randomly chosen from the remaining possible values. The features that are constant or randomly changing in the sequence are also constant or change randomly in all optional choice images.


\subsection{Abstract relations models}
\label{sec:model_comparisons}

\begin{figure*}
\begin{center}
\centerline{\includegraphics[width=0.8\linewidth]{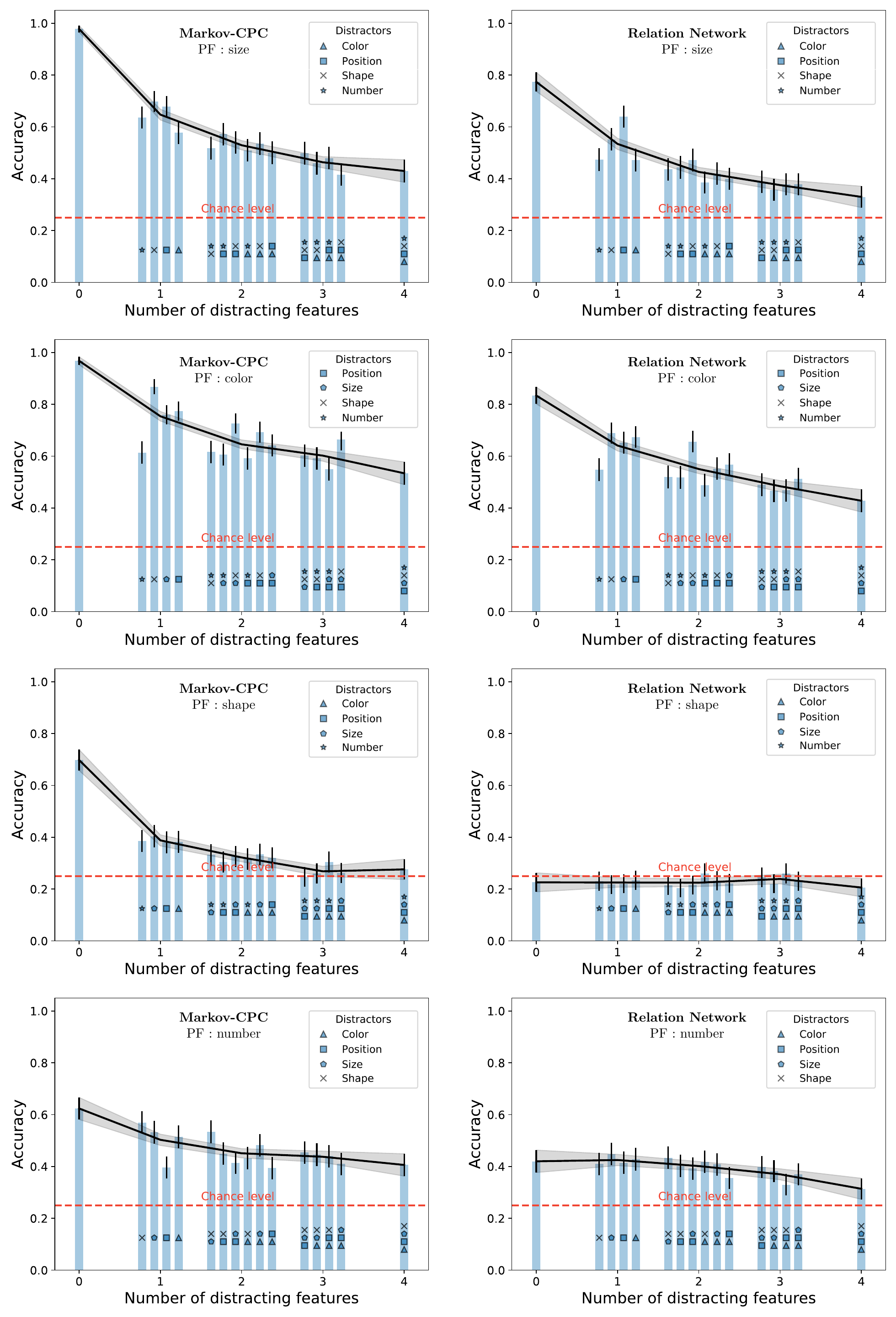}}
\caption{\textbf{Performance of the Markov-CPC (left column) and RN (right column) models on SCE tests with four different predictive features (rows).} For each predictive feature, we tested the networks over 16 test conditions where the rest of the features were either distractors (marked according to the legend) or constant (not marked). Each test condition included $500$ randomly generated intelligence tests. Error bars are 95\% confidence intervals. The black line and its shade are the average accuracy per difficulty and the standard deviation. The dashed line denotes the chance level ($\frac{1}{n}$).}
\label{fig:RN_vs_MCPC1D}
\end{center}
\end{figure*}

Images in the SCE task are related to each other by abstract relations. These abstract relations are between low-dimensional latent representations of the images. For example, two images $\mathbf{x_i}$ and $\mathbf{x_j}$ with the same number of shapes are related by their latent variables $Z(\mathbf{x}_i)$ and $Z(\mathbf{x}_j)$, which encode their number of shapes. Specifically, in their case, $Z(\mathbf{x}_i) = Z(\mathbf{x}_j)$. To account for relations more complex than equality and allow for these relations to be learned empirically, we used a learnable abstract relation function between images $\mathbf{x}_i$ and $\mathbf{x}_j$,
\begin{equation*}
    R_\theta\big(Z_\phi(\mathbf{x}_i), Z_\phi(\mathbf{x}_j)\big),
\end{equation*}
where $\theta$ and $\phi$ are the parameters of artificial neural networks $R$ and $Z$, respectively. 

We chose to construct a network that can identify the relation between consecutive sequence images in the sense that it can correctly identify the $K+1$ image out of the $n$ choice images. We compared two main candidates of deep learning models: Markov Contrastive Predictive Coding (Markov-CPC) \cite{oord_representation_2018} and Relation Network (RN) \cite{sung_learning_2018}. 

\subsubsection{Markov Contrastive Predictive Coding}

The Markov-CPC model (Fig. \ref{fig:two_models} left) has an inductive bias that assumes a causal predictive structure between inputs. It, therefore, uses a predictor function $T_\theta$ to predict a latent variable $Z_\phi(\mathbf{x}_j)$ given another latent variable $Z_\phi(\mathbf{x}_i)$. This manifests in the prediction error between the two variables, defined as,
\begin{equation*}
\label{eq:prediction_error}
\epsilon_{i,j}\left(Z_\phi,T_\theta\right)=\bigg(T_\theta\big(Z_\phi(\mathbf{x}_i)\big)-Z_\phi(\mathbf{x}_j)\bigg)^2.
\end{equation*}

To solve an SCE test, we used the model to reduce the prediction error between the latent variables of consecutive inputs. By construction, for an encoder and predictor functions that match the generative function and the deterministic rule of the SCE test, namely $Z^*=G^{-1}$ and $T^*=U$, the prediction error $\epsilon_{i,j}\left(Z^*,T^*\right)=0$ if $i$ and $j$ are two consecutive images ($j=i+1$), and $\epsilon_{i,j}\left(Z^*,T^*\right)>0$ otherwise. 

The challenge is that $Z^*$ and $T^*$ are unknown. However, given a sequence of $K$ ordered images, we can approximate $Z^*$ and $T^*$ by finding parameters $\phi$ and $\theta$ that minimize the prediction error for consecutive images and maximize it for the non-consecutive ones. For that, we defined a contrastive infoNCE loss based on those prediction errors,
\begin{equation*}
\label{eq:DNNs_CPC}
    \mathcal{L}_{M-CPC}=-\frac{1}{K-1}\sum_{i=1}^{K-1}\log{
    \frac{e^{-\epsilon_{i,i+1}}}
    {\sum_{i'=1}^{K}{e^{-\epsilon_{i,i'}}}}}
\end{equation*}
and find $\phi$ and $\theta$ that minimize it.

\subsubsection{Relation Network}
Unlike the Markov-CPC model, which is inductively biased for finding causal relations between images, the RN model (Fig. \ref{fig:two_models} right) can learn \textit{general} abstract relations. It achieves that by learning a general function $R_\theta\big(Z_\phi(x_i) \oplus Z_\phi(x_j)\big)$ where $\oplus$ is a concatenation operator.

Specifically for our tests, we used RN to learn the relation between consecutive pairs of inputs. Following \cite{sung_learning_2018}, we did that by using an MSE classification loss that classified consecutive and non-consecutive images. Given a sequence of $K$ images, we used the following RN loss function,
\begin{equation*}
\label{eq:DNNs_RN}
    \mathcal{L}_{RN}=\frac{1}{(K-1)^2}\sum_{i,j=1}^{K-1} \Big(R_\theta\big(Z_\phi(x_i) \oplus Z_\phi(x_j)\big) - \delta_{j,i+1} \Big)^2
\end{equation*}
in which the relation between consecutive images is assigned the label $1$, and between non-consecutive images - the label $0$.

\begin{table*}[b!]
 \small
  \centering
  \begin{tabular}{||l||c|c|c|c|c|c|c|c||}
    \hline
    \noalign{\vskip 0.5ex} \backslashbox{Training}{Testing} & \thead{Size\\(easy)} & \thead{Size\\(hard)}  & \thead{Color\\(easy)} &  \thead{Color\\(hard)} & \thead{Number\\(easy)} & \thead{Number\\(hard)} & \thead{Shape\\(easy)}  & \thead{Shape\\(hard)}\\ \noalign{\vskip 0.5ex}
    \hline   
    \noalign{\vskip 0.5ex} \thead{Size\\(easy)} & \makecell{$0.99$ \\ $\pm0.01$ }&\makecell{$0.97$ \\ $\pm0.01$ }&\makecell{$0.65$ \\ $\pm0.03$ }&\makecell{$0.68$ \\ $\pm0.03$ }&\makecell{$0.52$ \\ $\pm0.05$ }&\makecell{$0.5$ \\ $\pm0.03$ }&\makecell{$0.72$ \\ $\pm0.09$ }&\makecell{$0.42$ \\ $\pm0.14$ } \\
    \thead{Size\\(hard)} & \makecell{$0.9$ \\ $\pm0.0$ }&\makecell{$0.88$ \\ $\pm0.01$ }&\makecell{$0.24$ \\ $\pm0.0$ }&\makecell{$0.25$ \\ $\pm0.01$ }&\makecell{$0.24$ \\ $\pm0.01$ }&\makecell{$0.24$ \\ $\pm0.01$ }&\makecell{$0.32$ \\ $\pm0.04$ }&\makecell{$0.31$ \\ $\pm0.04$ }\\
    \thead{Color\\(easy)} & \makecell{$0.75$ \\ $\pm0.03$ }&\makecell{$0.78$ \\ $\pm0.01$ }&\makecell{$1.0$ \\ $\pm0.0$ }&\makecell{$1.0$ \\ $\pm0.0$ }&\makecell{$0.79$ \\ $\pm0.05$ }&\makecell{$0.81$ \\ $\pm0.05$ }&\makecell{$0.61$ \\ $\pm0.13$ }&\makecell{$0.71$ \\ $\pm0.1$ }\\
    \thead{Color\\(hard)} & \makecell{$0.26$ \\ $\pm0.01$ }&\makecell{$0.27$ \\ $\pm0.01$ }&\makecell{$0.93$ \\ $\pm0.01$ }&\makecell{$0.95$ \\ $\pm0.01$ }&\makecell{$0.27$ \\ $\pm0.01$ }&\makecell{$0.26$ \\ $\pm0.01$ }&\makecell{$0.53$ \\ $\pm0.03$ }&\makecell{$0.34$ \\ $\pm0.05$ }\\
    \thead{Number\\(easy)} & \makecell{$0.45$ \\ $\pm0.01$ }&\makecell{$0.44$ \\ $\pm0.01$ }&\makecell{$0.41$ \\ $\pm0.01$ }&\makecell{$0.39$ \\ $\pm0.02$ }&\makecell{$0.67$ \\ $\pm0.03$ }&\makecell{$0.6$ \\ $\pm0.03$ }&\makecell{$0.33$ \\ $\pm0.06$ }&\makecell{$0.32$ \\ $\pm0.07$ }\\
    \thead{Number\\(hard)} & \makecell{$0.26$ \\ $\pm0.0$ }&\makecell{$0.26$ \\ $\pm0.01$ }&\makecell{$0.28$ \\ $\pm0.01$ }&\makecell{$0.25$ \\ $\pm0.01$ }&\makecell{$0.7$ \\ $\pm0.02$ }&\makecell{$0.67$ \\ $\pm0.03$ }&\makecell{$0.33$ \\ $\pm0.03$ }&\makecell{$0.26$ \\ $\pm0.04$ }\\
    \thead{Shape\\(easy)} & \makecell{$0.18$ \\ $\pm0.02$ }&\makecell{$0.25$ \\ $\pm0.03$ }&\makecell{$0.29$ \\ $\pm0.02$ }&\makecell{$0.33$ \\ $\pm0.03$ }&\makecell{$0.2$ \\ $\pm0.02$ }&\makecell{$0.26$ \\ $\pm0.03$ }&\makecell{$0.37$ \\ $\pm0.06$ }&\makecell{$0.31$ \\ $\pm0.05$ }\\
    \thead{Shape\\(hard)} & \makecell{$0.24$ \\ $\pm0.01$ }&\makecell{$0.25$ \\ $\pm0.01$ }&\makecell{$0.26$ \\ $\pm0.01$ }&\makecell{$0.26$ \\ $\pm0.0$ }&\makecell{$0.24$ \\ $\pm0.01$ }&\makecell{$0.25$ \\ $\pm0.01$ }&\makecell{$0.27$ \\ $\pm0.01$ }&\makecell{$0.24$ \\ $\pm0.02$ }\\
    
    \thead{Naive} & \makecell{$0.97$\\ $\pm0.0$}& \makecell{$0.42$\\ $\pm0.01$ }& \makecell{$0.96$\\ $\pm0.0$ }& \makecell{$0.55$\\ $\pm0.01$} & \makecell{  $0.60$\\ $\pm0.01$} &\makecell{ $0.40$\\ $\pm0.01$} & \makecell{$0.71$\\ $\pm0.0$ }& \makecell{$0.26$\\ $\pm0.01$}\\
    \hline
  \end{tabular}
  \vskip 1ex
  \caption{Each cell in the table is the average accuracies of 10 networks, each trained on 1000 episodes of one rule and tested using 500 tests on a different (or the same) rule. The errors are 95\% confidence intervals.}
  \label{table:cross_domain}
\end{table*}

\section{Results}
\subsection{Naive few-shot learning}

We applied the Markov-CPC and RN models to an SCE test in the following way: First, we randomly initialized the models' networks (see architectures in Appendix \ref{sec:model}). We then updated these networks' weights with a \textit{single} optimization step in the direction that minimizes the RN or Markov-CPC loss functions, given the $K=5$ sequence images (optimizer details are also in Appendix \ref{sec:model}). After the single optimization step, we evaluated the consistency of each choice image with the sequence based on the resulting Markov-CPC or RN loss function, when these choices were applied as the sixth image. We selected the most consistent choice image, out of the $n=4$ choices, as the answer.

We found that both models performed much better than chance ($0.25$) in almost all tasks and all levels of difficulty (Fig. \ref{fig:RN_vs_MCPC1D}). However, their performance decreased with the number of distractors, indicating that this number is a good measure of the task's difficulty. We also found that some rules, Color and Size, seem easier than other rules. For example, Markov-CPC perfectly performed the easier Size and Color tasks. On the other hand, the Number rule seemed more difficult, which indicates that numerosity is not encoded accurately by naive networks. Notably, a similar claim has been made in the psychology literature about the encoding of numbers in ``naive'' humans \cite{leibovich_katzin_harel_henik_2017, 10.3389/fpsyg.2018.00571}. Overall, the Markov-CPC performed 
better than the RN. This was particularly pronounced when considering the Shape rule, in which the performance of RN, unlike Markov-CPC, did not exceed the chance level. Importantly, naive networks that did not train on the sequence images (did not perform the single gradient step) did not solve the task (see Appendix \ref{sec:control}).

To study the determinants of the model's performance, we tested various variations of Markov-CPC and RN (see Appendix \ref{sec:other_variants}). The emerging picture is that a match between the inductive bias of the model and the task crucially affects its performance. First, the Markov CPC that posits a causal relation between the consecutive images does better than the RN model, which allows for a more general relation between the images. Second, in our task, which is Markovian, the performance of a non-Markovian CPC, which allows a more complex relation between the latent variables, is slightly worse than that of the more restrictive Markov-CPC. Third, using the particular residual predictor $T_\theta$ ($T_\theta(Z)=Z+\Delta T_\theta(Z)$) resulted in higher performance than a non-residual predictor $T_\theta$. Finally, increasing the dimensionality of the latent variable was detrimental to performance (albeit the effect was small). 

Additionally, adding complexity to the model did not improve its performance: both for the Markov CPC and the RN models, the relatively shallow network we used for the encoder $Z$ was better than deeper and more complicated ones. This result may seem contradictory to the general trend of preferring ever-deeper networks. However, achieving higher performance in these more complex networks requires more data, while our networks were trained on a minimal number of examples. 

\begin{figure*}[b!]
\begin{center}
\includegraphics[width=0.8\linewidth]{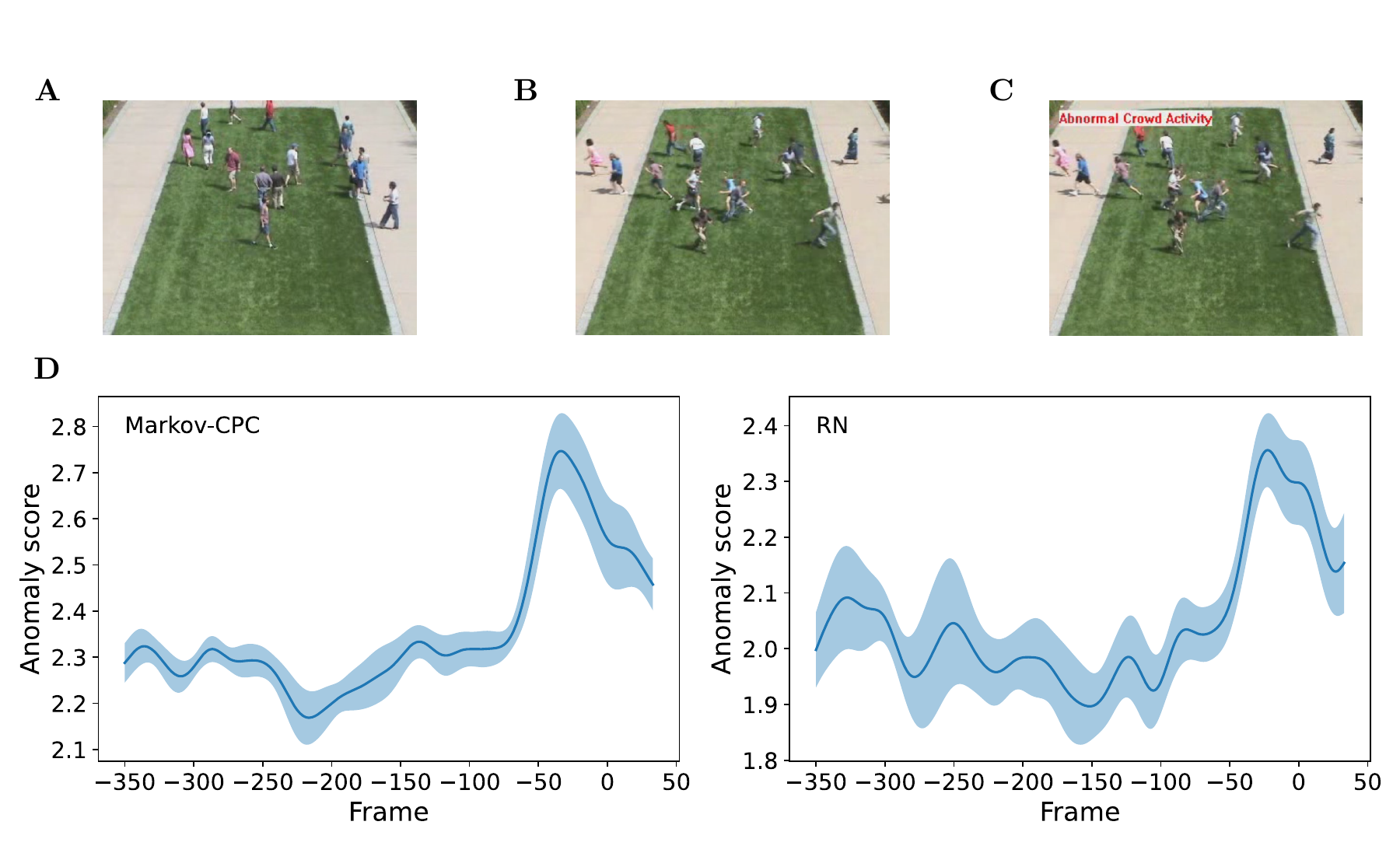}
\caption{\textbf{UMN Anomaly scores.} (A-C) Example frames from the first UMN video: (A) normal walking human behavior; (B) humans start to run; (C) “Abnormal Crowd Activity” label appears. (D) Anomaly scores of the frames. The frames are numbered with respect to the first appearance of the “Abnormal Crowd Activity” label. We ran a model 5 times for each frame. We averaged the anomaly scores over the 5 runs per frame and smoothed the scores with a Gaussian kernel with a standard deviation of 10 frames. Finally, we averaged the resulting scores over the 11 videos.}
\label{fig:anomaly_detection2}
\end{center}
\end{figure*}

\subsection{Prior training}
\label{sec:prior_training}
\subsubsection{Expressivity}

While Markov-CPC performed substantially better than chance, it still failed to select the correct choice image in some tests. To see whether this resulted from a limited expressivity of the model, we pretrained Markov-CPC with SCE tests as training episodes. Specifically, we performed one optimization step for each of these training-episode tests and then tested the model on novel tests. We found that with $1000$ training episodes, the performance of Markov-CPC on tests in which Size and Color are the predictive features can exceed 90\%, even in the hardest trials (Table \ref{table:cross_domain}, best performance for each testing rule). These results indicate that, at least for these predictive features, the performance of the Markov-CPC model is not limited by its expressivity. By contrast, when the Number or Shape were the predictive features, performance level improved with training but could only reach 70\%--80\%, leaving the question of expressivity of the model open.

\subsubsection{Specialization}
\label{sec:specialization}

The challenge of fluid intelligence is identifying regularities in a domain on which the agent has not been trained. One possibility to address this problem is to train on a large number of different examples, with the hope that they will generalize to the new domain. To test whether our networks can generalize, we tested whether pretraining in one type of test improves (or impairs) the model's performance in another type of test. We considered 8 different test types for training episodes (four different predictive features, either in the easiest setting of no distractors or in the hardest setting of four distractors). We tested the performance on those 8 different tests, yielding an $8\times8$ performance matrix (Table \ref{table:cross_domain}). We found that typically, pretraining within a domain (same predictive feature) improved performance in that domain relative to the naive network. The exception was training on Shape, which, surprisingly, was detrimental to performance on Shape tests. Interestingly, we noted that within a domain, training using easy episodes was typically more effective than training using hard ones. 

However, training in one domain was typically detrimental to performance in other domains (relative to naive networks). This result highlights the potential advantage of ``fluid'' models over trained ones when the test domain is unknown.

\begin{figure*}[t!]
\begin{center}
\includegraphics[width=0.8\linewidth]{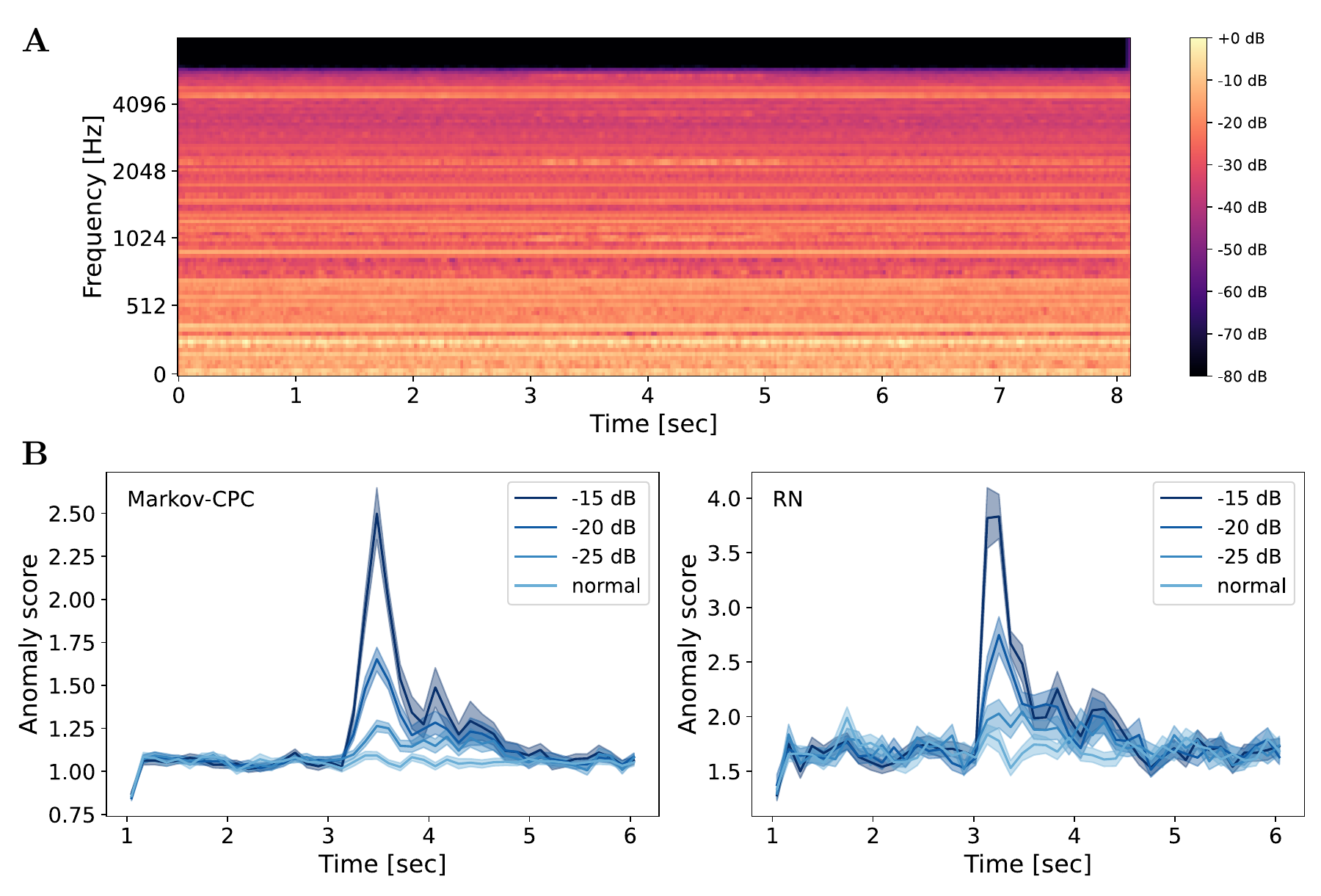}
\vskip -0.05in
\caption{\textbf{ADS anomaly scores.} (A) Example spectrogram of a phone that rings over natural background noise. The anomaly-to-noise (ANR) ratio in this example is -20 dB. (B) Anomaly scores of Markov-CPC and RN. We averaged the anomaly scores over five iterations per frame of each sound snippet and the 140 sound snippets in each ANR.}
\label{fig:anomaly_detection3}
\end{center}
\vskip -0.1in
\end{figure*}

\subsection{Anomaly detection}
\label{sec:anomaly_detection}

To test naive few-shot learning models in natural settings and compare the performance of Markov-CPC and RN, we tested the abilities of these models to detect unlabeled anomalies in two datasets. 

The first dataset, UMN \cite{mehran_abnormal_2009}, consists of 11 security camera videos that A) start with humans walking normally; B) towards the end of the videos, they start running; C) a label of "Abnormal Crowd Activity" appears shortly after\footnote{In the tests we removed the top 30 pixels of all images to exclude this label.} (Fig. \ref{fig:anomaly_detection2} A-C). We assigned each video frame an anomaly score based on its consistency with its five preceding frames (see details in appendix \ref{sec:anomaly_detection_appendix}). We found that both the Markov-CPC and the RN models assigned larger anomaly scores to the frames in which the humans started to run (Fig. \ref{fig:anomaly_detection2}). Detecting the time in which running begins is not a difficult task for trained networks, and previous anomaly detection models achieved nearly perfect scores in this dataset \cite{pang_self-trained_2020}. However, these previous models all relied on prior training. By contrast, our objective was to identify these anomalies without prior training. 

\begin{figure*}[t!]
\begin{center}
\includegraphics[width=\linewidth]{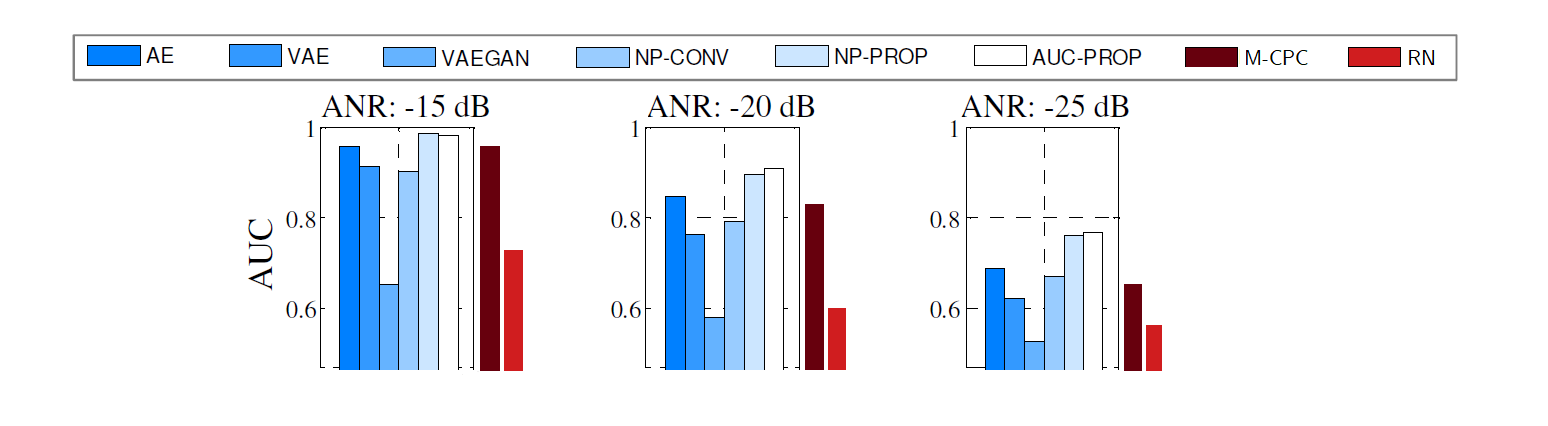}
\caption{\textbf{ADS model comparison.} AUC scores for anomaly detection of different models. Bars in the boxes depict the AUC scores of six models that require prior training (adapted from \cite{koizumi_unsupervised_2019}). Darker and lighter red depict the AUC scores of the Markov-CPC and RN models, respectively.}
\label{fig:anomaly_detection4}
\end{center}
\vskip -0.1in
\end{figure*}

The second dataset, Anomaly Detection of Sound (ADS) \cite{koizumi_unsupervised_2019}, consists of 140 short sound snippets. Each snippet consists of natural background noise, such as the noise of an air conditioner recorded in a natural environment, interrupted by various types of anomalous sounds, such as keys falling or a drawer closing. The detection difficulty was determined by the anomaly-to-noise power ratios (ANRs), which were set to -15 dB, -20 dB, or -25 dB. This dataset is more challenging than the UMN dataset; even trained networks fail to detect anomalous sounds in some examples. To test our models on this dataset, we converted the sound snippets into a video in which each frame was a Mel-spectrogram window of $\sim 0.5$ seconds of the snippet, hopped in $\sim 0.1$ second steps. Then, similar to the analysis of the UMN dataset, we computed an anomaly score for each image based on the previous five frames using either the Markov-CPC or the RN models. We found that both models assigned relatively higher anomaly scores to anomalous events (Fig. \ref{fig:anomaly_detection3}). 

To compare our results with other models, we used the standard AUC measure (area under the ROC curve) \cite{pedregosa2011scikit} based on the maximal anomaly score of each snippet. For Markov-CPC, The AUC scores were around 0.96, 0.84, and 0.67 for the -15, -20, and -25 dB ANRs respectively, which are not much worse than those of previous models. This is despite the fact that our model was naive, whereas previous models required substantial prior training (Fig. \ref{fig:anomaly_detection4}). The RN model exhibited substantially poorer performance in this task than the Markov-CPC model. The finding that the Markov-CPC model performs better than the RN model is consistent with its relative success in the SCE tests. 

\section{Discussion}

Our results showed that deep learning tools could be used to solve non-trivial tasks without any prior training. The Markov-CPC model successfully solved SCE tests and detected anomalies in data streams, starting with random weights. This approach has practical implications. Today, deep learning models lack ``fluid'' abilities, impairing their performance in uncertain and shifting environments such as driving in bad weather conditions \cite{AdveWeather}. Markov-CPC's ability to detect anomalies in data streams without prior training is useful in those uncertain environments, where prior training might even be detrimental (Section \ref{sec:specialization}).

Moreover, our work suggests a method to develop fluid intelligent models. A complex task can be separated into sub-components. Our findings suggest that deep learning models can solve some of these sub-components without relying on prior knowledge. Like our anomaly detection model, these fluid-intelligence models will be flexible to changing environments. The main challenge is finding these sub-components for which there is an appropriate inductive bias. We encourage the finding of such sub-components, and the further improvement of naive models' ability to solve the SCE task.

Regarding the relationship between our approach and standard deep learning methodology, in which models are extensively trained to solve complex problems. While we focused on naive untrained models, it is, of course, possible to pretrain fluid intelligent models. Indeed, today's standard few-shot learning approach is to pretrain deep learning models over large relevant, diverse datasets and then fine-tune them for the new tasks \cite{brown_language_2020, reed_generalist_2022}. This approach has limited generalization ability to datasets that are different from the datasets they were trained on \cite{li_learning_2017, nalisnick_deep_2019, yin_meta-learning_2020,  rajendran_meta-learning_2020}. However, we expect the combination of pretrained weights with high fluid intelligence to work best in relatively stable environments.

In a field that is dominated by extensive training with an extensive number of examples, our work demonstrates the potential power of ``fluid'' models. Pursuing this line of research, making artificial fluid-intelligent models, will make machines considerably more flexible.


\bibliography{ICML_2023.bib}
\bibliographystyle{icml2023}

\newpage
\appendix
\onecolumn

\renewcommand{\thesection}{A\arabic{section}}
\renewcommand{\thefigure}{f\arabic{figure}}
\renewcommand{\thetable}{t\arabic{table}}
\setcounter{figure}{0}
\setcounter{table}{0}
\section*{Appendix}

\section{Models details}
\label{sec:model}

\subsection{Markov-CPC}

Markov-CPC's encoder $Z$ and predictor $T$ were implemented by deep neural networks. We used a relatively shallow convolutional neural network for the encoder $Z$. For the predictor $T$, we used a residual network, such that \[T\big(Z(\mathbf{x})\big)=Z(\mathbf{x}) + \Delta T\big(Z(\mathbf{x})\big)  \] where $\Delta T$ is a fully connected neural network.

\begin{figure*}[h!]
\begin{center}
\includegraphics[width=0.7\linewidth]{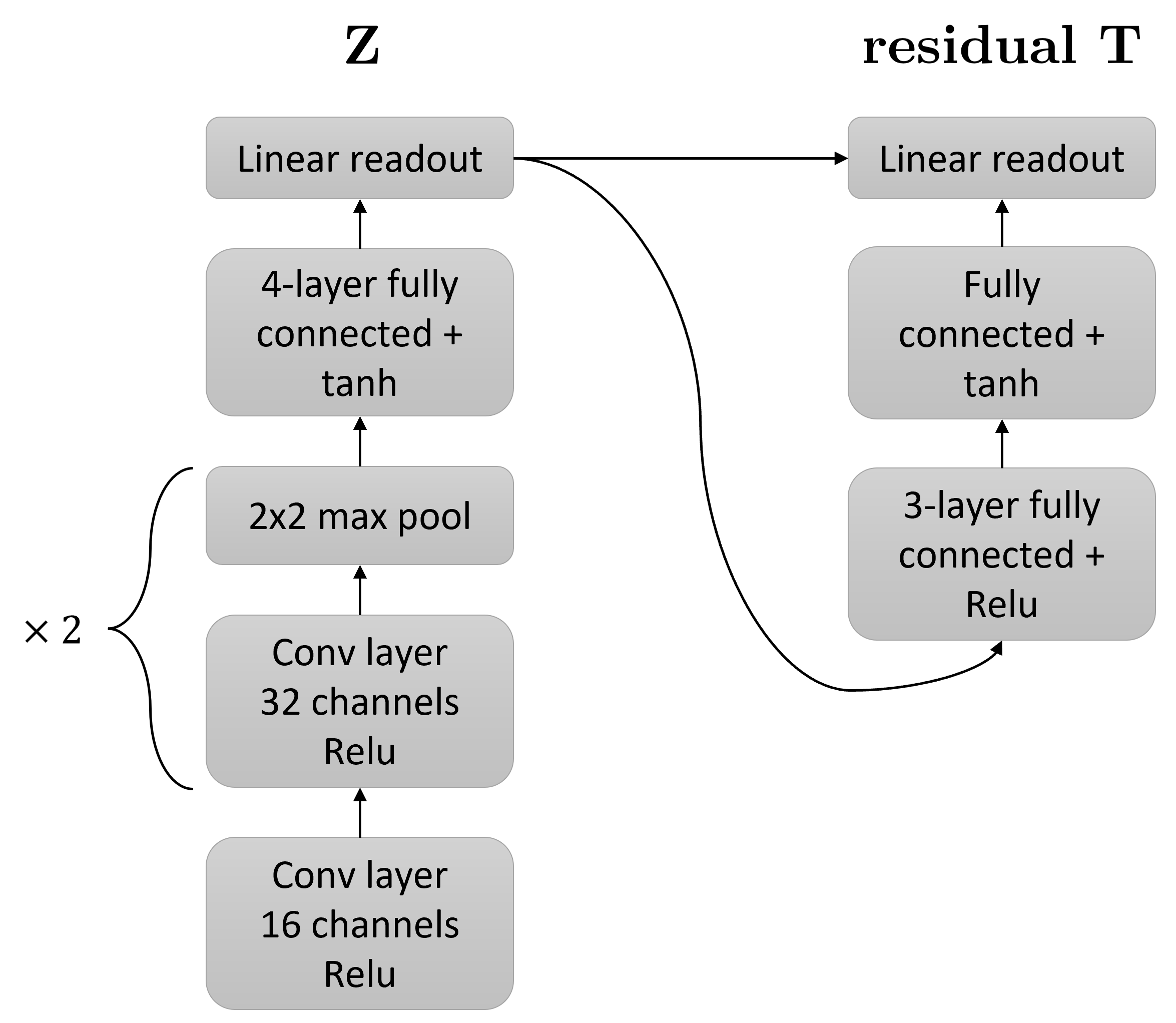}
\caption{\textbf{M-CPC architecture.}}
\label{fig:M-CPC_architecture}
\end{center}
\end{figure*}

To update the weights of the networks, we used the RMSprop optimizer with a learning rate $\eta=4\cdot10^{-4}$. The rest of the optimizer hyperparameters were set to PyTorch defaults.

\subsection{Relation Network}

We used the same Relation Network loss function as in the original paper \cite{sung_learning_2018}. For the networks, we either used the networks from \cite{sung_learning_2018} or used the same $Z$ and $R$ as Markov-CPC's $Z$ and $T$. In both cases, we used the RMSprop optimizer. For the original RN networks from \cite{sung_learning_2018} we used a learning of $\eta=4\cdot10^{-6}$. When we used the $Z$ of Markov-CPC, we used the learning rate $\eta=4\cdot10^{-4}$. The rest of the optimizer parameters were set to PyTorch default.

\newpage

\section{Control experiment}
\label{sec:control}

\begin{figure*}[b!]
\vskip -0.2in
\begin{center}
\centerline{\includegraphics[width=0.8\linewidth]{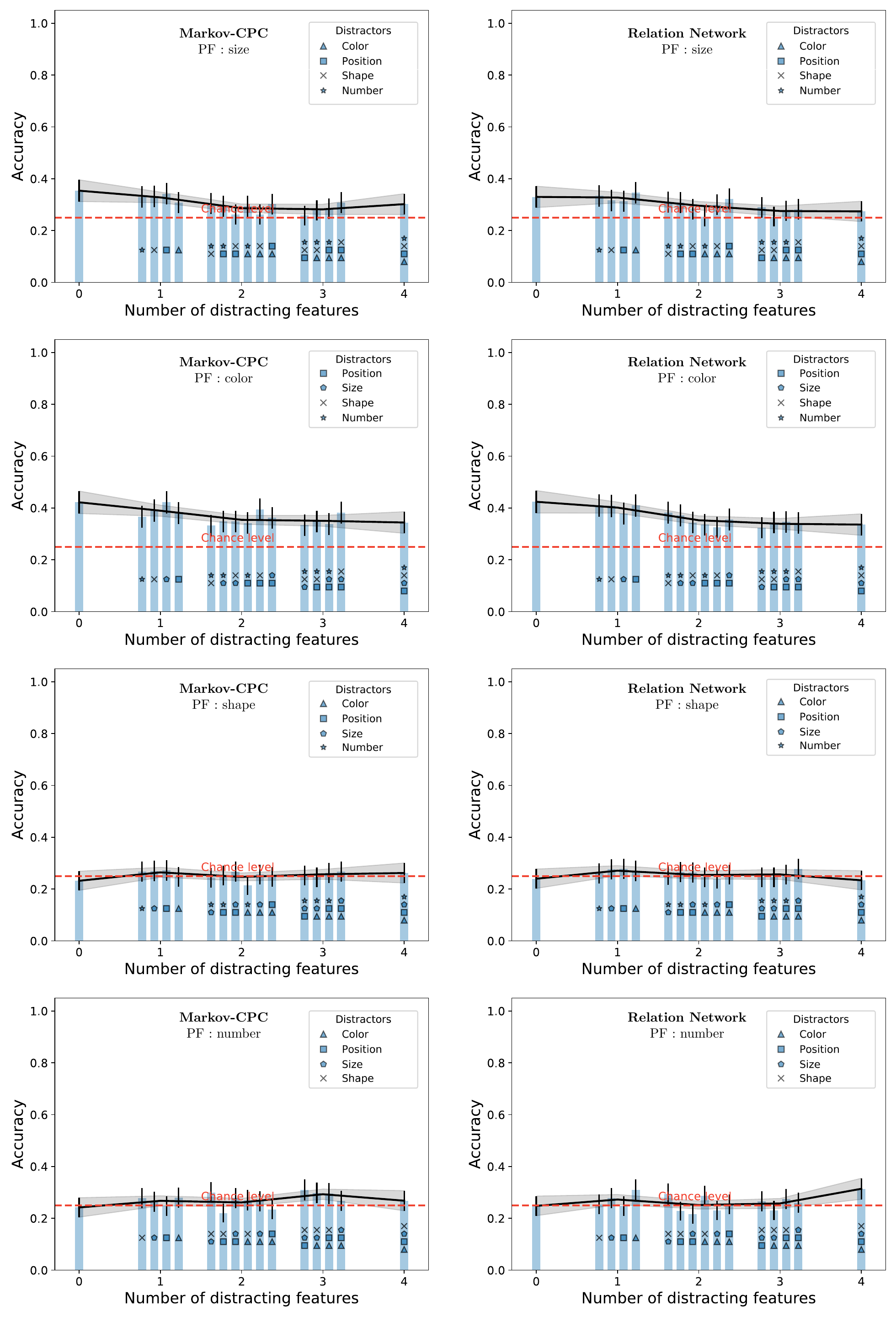}}
\caption{\textbf{Control experiment.} The accuracies that Markov-CPC and the RN models achieved when they solved tests without performing a gradient step.}
\label{fig:M-CPC_architecture}
\end{center}
\end{figure*}

\newpage

\section{Anomaly score based on SCE}
\label{sec:anomaly_detection_appendix}



To assign an anomaly score to a candidate image $\mathbf{x}_c$ based on its preceding images, we optimized a naive Markov-CPC or RN model, with a single optimization step, on the $K=5$ preceding images. To determine the candidate's anomaly score $s_c$ with the Markov-CPC model, we checked by how much the prediction error of the candidate image given the $K^{\text{th}}$ image, $\epsilon_{K,c}$, deviates from the prediction errors of the $K-1$ preceding consecutive pair images. Mathematically, the score is defined by,
\begin{equation*}
    s_c = \frac{\epsilon_{K,c}-\langle\epsilon_p\rangle}{\text{std}\left(\epsilon_p\right)},
\end{equation*}
where $\langle\epsilon_p\rangle$ is the average prediction error of the $K-1$ preceding consecutive pairs and $\text{std}\left(\epsilon_p\right)$ is their standard deviation.

To determine $s_c$ with the RN model, we checked by how much the candidate's classification error, 
\begin{equation}
\label{RN_class_error}
    \epsilon_{K,c}=\Big(R_\theta\big(Z_\phi(x_K) \oplus Z_\phi(x_c)\big) - 1 \Big)^2
\end{equation}
deviates from the classification errors of the $K-1$ preceding consecutive pair images. We used the same mathematical score defined above, where the classification errors are used to define as in Eq (\ref{RN_class_error}).

\section{Model variations}
\label{sec:other_variants}

Markov-CPC achieved an average accuracy of $0.52\pm0.02$ over all the test conditions, while RN's average accuracy was $0.42\pm0.02$. We tested how this average accuracy changes for various model variations.

Markov-CPC was implemented with a residual $T$, such that $T\big(Z(\mathbf{x})\big)=Z(\mathbf{x})+\Delta T\big(Z(\mathbf{x})\big)$ where $\Delta T$ is a neural network (see appendix \ref{sec:model}). we also tested a variant of Markov-CPC in which $T$ is non-residual. We found that the performance of a Markov-CPC with a non-residual $T$ is substantially worse than that with a residual $T$ (Table \ref{table:residual_vs_not}). This indicates that a residual $T$ is a good inductive bias for finding these rules. 

\begin{table}[H]
  \centering
  \begin{tabular}{||l|c||}
    \hline
    \noalign{\vskip 0.5ex} \textbf{Model variant} & \textbf{Total accuracy}      \\ 
    \hline   
    \noalign{\vskip 0.5ex} \textbf{Residual $T$} & $\mathbf{0.52\pm0.02}$ \\
    Non-residual $T$ & $0.29\pm0.02$ \\
    \hline
  \end{tabular}
    \vskip 1ex
  \caption{Residual versus non-residual $T$.}
  \label{table:residual_vs_not}
\end{table}

CPC models are usually non-Markovian \cite{oord_representation_2018,henaff_data-efficient_2020}. To test the effect of memory on the performance of the CPC model, we used latent variables $Z$ whose values were also dependent on previous images in the sequence, either via a regular recurrent neural network (RNN) or an LSTM \cite{hochreiter_long_1997}. Both the recurrent connections and the LSTM impaired performance, while taking twice the time to compute (Table \ref{table:markov_rnns}). This indicates that adding recurrent weights to the latent variables, when the data can be explained by Markov latent variables, can impair the ability of the network to extract the rule.

\begin{table}[H]
  \centering
  \begin{tabular}{||l|c|c||}
    \hline
    \noalign{\vskip 0.5ex} \textbf{Model variant} & \textbf{Total accuracy} & \textbf{Tests per second}     \\ 
    \hline   
    \noalign{\vskip 0.5ex} \textbf{Markov-CPC} & $\mathbf{0.52\pm0.02}$ & $\mathbf{13 \pm 2}$  \textbf{[Hz]}\\
    LSTM-CPC & $0.48\pm0.02$ & $6.6 \pm 0.4$  [Hz]\\ 
    RNN-CPC & $0.47\pm0.02$ & $6.6 \pm 0.4$ [Hz] \\ 
    \hline
  \end{tabular}
      \vskip 1ex
  \caption{Markov versus recurrent networks. The running times were measured on a basic laptop with Nvidia RTX 2070 GPU.}
  \label{table:markov_rnns}
\end{table}

Throughout the paper, we used Markov-CPC with a 1-dimensional latent space. We tried changing the latent space dimension of Markov-CPC. We found that even a 1-dimensional latent space is enough and that the performance does not strongly depend on this value (Table \ref{table:Z_dim}). 

\begin{table}[H]
  
  \centering
  \begin{tabular}{||l|c||}
    \hline
    \noalign{\vskip 0.5ex} \textbf{Model variant} & \textbf{Total accuracy}      \\ 
    \hline   
    \noalign{\vskip 0.5ex} Markov-CPC-1D & $0.52\pm0.02$ \\
    Markov-CPC-10D & $0.53\pm0.02$  \\
    Markov-CPC-100D & $0.51\pm0.02$  \\
    Markov-CPC-1000D & $0.50\pm0.02$  \\
    \hline
  \end{tabular}
    \vskip 1ex
  \caption{Different $Z$ latent dimensions.}
  \label{table:Z_dim}
\end{table}



We also measured the performance of the model without a contrastive loss, minimizing the prediction errors (Eq. \ref{eq:prediction_error}) of consecutive inputs only (Table \ref{table:no_contrast}).

\begin{table}[H]
  \centering
  \begin{tabular}{||l|c||}
    \hline
    \noalign{\vskip 0.5ex} \textbf{Model variant} & \textbf{Total accuracy}      \\ 
    \hline   
    \noalign{\vskip 0.5ex} \textbf{Contrastive loss} & $\mathbf{0.52\pm0.02}$ \\
    No contrast & $0.45\pm0.02$ \\
    \hline
  \end{tabular}
    \vskip 1ex
    \caption{No contrast.}
  \label{table:no_contrast}
\end{table}

We trained Markov-CPC throughout the paper with an RMSprop optimizer. We also measured the performance with a standard SGD optimizer (Table \ref{table:SGD}). The optimal learning rate was found to be $\eta=40$.

\begin{table}[H]
  \centering
  \begin{tabular}{||l|c||}
    \hline
    \noalign{\vskip 0.5ex} \textbf{Model variant} & \textbf{Total accuracy}      \\ 
    \hline   
    \noalign{\vskip 0.5ex} RMSprop, $\text{lr}=4\cdot10^{-4}$ & $0.52\pm0.02$ \\
    SGD,  $\text{lr}=40$ & $0.5\pm0.02$ \\
    \hline
  \end{tabular}
  \vskip 1ex
   \caption{SGD.}
  \label{table:SGD}

\end{table}

We also compared the performance of the original RN networks \cite{sung_learning_2018} to the more shallow networks that we used for Markov-CPC (Fig. \ref{fig:M-CPC_architecture}). The shallower networks were better (Table \ref{table:shallow_vs_deep_RN}), indicating that in the RN models, simpler networks are better in this task than deeper networks.

\begin{table}[H]
  \centering
  \begin{tabular}{||l|c||}
    \hline
    \noalign{\vskip 0.5ex} \textbf{Model variant} & \textbf{Total accuracy}      \\ 
    \hline  
    \noalign{\vskip 0.5ex} \textbf{Shallow RN (Fig. \ref{fig:M-CPC_architecture})} & $\mathbf{0.42\pm0.02}$ \\
    Deep RN \cite{sung_learning_2018} & $0.27\pm0.02$ \\
    \hline
  \end{tabular}
  \vskip 1ex
  \caption{Deep versus shallow RN networks.}
   \label{table:shallow_vs_deep_RN}
\end{table}

Both in the Markov-CPC and RN models, we used a relatively shallow network for the encoder Z (Fig. \ref{fig:M-CPC_architecture}). We tested other, more complex and deep, networks from the literature as candidate encoder backbones of the Markov-CPC model. The shallow encoder was better than various complicated and deep networks from the literature (Table \ref{table:backbones}).

\begin{table}[H]
  \centering
  \begin{tabular}{||l|c||}
    \hline
    \noalign{\vskip 0.5ex} \textbf{Model variant} & \textbf{Total accuracy}      \\ 
    \hline  
    \noalign{\vskip 0.5ex} \textbf{Paper's network (Fig. \ref{fig:M-CPC_architecture})} & $\mathbf{0.52\pm0.02}$ \\
    VGG11 \cite{simonyan2014very} &$0.45\pm0.02$* \\
    DenseNet121 \cite{huang2017densely} & $0.33\pm0.02$ \\
    AlexNet \cite{NIPS2012_c399862d} & $0.31\pm0.02$ \\
    ResNet18 \cite{he2016deep} & $0.29\pm0.02$* \\
    MobileNet v3 small \cite{howard2017mobilenets} & $0.29\pm0.02$ \\
    \hline
  \end{tabular}
  \vskip 1ex
  \caption{Encoder Z networks comparison. *ResNet18 and VGG11 achieved relatively high accuracy on the color rule ($0.47\pm0.04$ and $0.78\pm0.04$ respectively).}
   \label{table:backbones}
\end{table}

\end{document}